\documentclass[a4paper, 10pt, conference]{ieeeconf} 

\IEEEoverridecommandlockouts 

\usepackage[keeplastbox]{flushend}
\usepackage{bm}

\usepackage[pdftex]{graphicx}
\usepackage{algorithmic}
\usepackage{mathptmx} 
\usepackage{amsmath} 
\usepackage{amssymb}  
\usepackage{subcaption}
\usepackage{url}
\usepackage{tabularx}
\usepackage{booktabs}
\usepackage{orcidlink}
\usepackage{multirow}
\usepackage[acronym,nonumberlist]{glossaries}
\usepackage[local, markifdraft]{gitinfo2}
\usepackage{dblfloatfix}    
\newglossaryentry{ADAS}{type=\acronymtype, name={ADAS}, description={Advanced Driver Assistance System}, first={Advanced Driver Assistance System (ADAS)}}
\newglossaryentry{ViL}{type=\acronymtype, name={ViL}, description={Vehicle-in-the-loop}, first={ViL}}
\newglossaryentry{VuT}{type=\acronymtype, name={VuT}, description={vehicle under test}, first={vehicle under test (VuT)}}
\newglossaryentry{HiL}{type=\acronymtype, name={HiL}, description={Hardware-in-the-loop}, first={HiL}}
\newglossaryentry{SiL}{type=\acronymtype, name={SiL}, description={Software-in-the-loop}, first={SiL}}
\newglossaryentry{MiL}{type=\acronymtype, name={MiL}, description={Model-in-the-loop}, first={MiL}}
\newglossaryentry{VEHiL}{type=\acronymtype, name={VEHiL}, description={Vehicle Hardware-in-the-loop}, first={VEHiL}}
\newglossaryentry{ACC}{type=\acronymtype, name={ACC}, description={Adaptive Cruise Control}, first={Adaptive Cruise Control (ACC)}}
\newglossaryentry{AV}{type=\acronymtype, name={AV}, description={Automated Vehicles}, first={Automated Vehicle (AV)}}
\newglossaryentry{TFNDS}{type=\acronymtype, name={TFNDS}, description={Test Bed
            Lower Saxony}, first={Test Bed Lower Saxony (TFNDS)}}
\newglossaryentry{DLR}{type=\acronymtype, name={DLR}, description={Germany Aerospace Center}, first={Germany Aerospace Center (DLR)}}
\newglossaryentry{ANN}{type=\acronymtype, name={ANN}, description={Artificial Neuronal Network}, first={Artificial Neuronal Network (ANN)}}
\newglossaryentry{DTW}{type=\acronymtype, name={DTW}, description={Dynamic Time Warping}, first={Dynamic Time Warping (DTW)}}
\newglossaryentry{SMoS}{type=\acronymtype, name={SMoS}, description={Surrogate Measure of Safety}, first={Surrogate Measure of Safety (SMoS)}, firstplural={Surrogate Measures of Safety (SMoS)}}

\newglossaryentry{RNN}{type=\acronymtype, name={RNN}, description={Recurrent Neuronal Network}, first={Recurrent Neuronal Network (RNN)}}

\newglossaryentry{SVM}{type=\acronymtype, name={SVM}, description={Support Vector Machine}, first={Support Vector Machine (SVM)}}
\newglossaryentry{DP}{type=\acronymtype, name={DP}, description={Driving primitive}, first={driving primitive (DP)}, firstplural={driving primitives (DP)}}
\newglossaryentry{EM}{type=\acronymtype, name={EM}, description={Expectation maximization}, first={expectation maximization (EM)}}

\newglossaryentry{DHC}{type=\acronymtype, name={DHC}, description={Divisive Hierarchical Clustering}, first={Divisive Hierarchical Clustering (DHC)}}
\newglossaryentry{AHC}{type=\acronymtype, name={AHC}, description={Agglomerative Hierarchical Clustering}, first={Agglomerative Hierarchical Clustering (AHC)}}

\newglossaryentry{HMM}{type=\acronymtype, name={HMM}, description={Hidden Markov Model}, first={Hidden Markov Model (HMM)}}

\newglossaryentry{sgm}{type=\acronymtype, name={SGM}, description={Semi-Global Matching}, first={Semi-Global Matching (SGM)}}
\newglossaryentry{ivs}{type=\acronymtype, name={IVS}, description={Intelligent Video-System}, first={Intelligent Video-System (IVS)}, plural={IVS}, firstplural={Intelligenten Video-Systemen(IVS)}}

\newglossaryentry{mog}{type=\acronymtype, name={MoG}, description={Mixture of Gaussian}, first={Mixture of Gaussian (MoG)}}

\newglossaryentry{GMM}{type=\acronymtype, name={GMM}, description={Gaussian Mixture Model}, first={Gaussian Mixture Model (GMM)}, firstplural={Gaussian Mixture Models (GMM)}, plural={Gaussian Mixture Models (GMM)}}
\newglossaryentry{gmmd}{type=\acronymtype, name={GMMD}, description={Gaussian Mixture Model using Distance}, first={Gaussian Mixture Model using Distance (GMMD)}}
\newglossaryentry{mf}{type=\acronymtype, name={MF}, description={Median Filter}, first={Median Filter (MD)}}
\newglossaryentry{fd}{type=\acronymtype, name={FD}, description={Frame Differencing}, first={Frame Differencing (FD)}}

\newglossaryentry{ptz}{type=\acronymtype, name={PTZ}, description={pan-tilt-zoom}, first={pan-tilt-zoom (PTZ)}}
\newglossaryentry{isg}{type=\acronymtype, name={ISG}, description={International Security Group}, first = {International Security Group (ISG)}}
\newglossaryentry{poe}{type=\acronymtype, name={PoE}, description={Power over Ethernet}, first = {Power over Ethernet (PoE)}}
\newglossaryentry{genicam}{type=\acronymtype, name={GenICam}, description={Generic Interface for Cameras}, first = {Generic Interface for Cameras (GenICam)}}
\newglossaryentry{xml}{type=\acronymtype, name={XML}, description={Extensible Markup Language}, first = {Extensible Markup Language (XML)}}
\newglossaryentry{pwm}{type=\acronymtype, name={PWM}, description={Pulsweitenmodulation}, first = {Pulsweitenmodulation (PWM)}}
\newglossaryentry{ncia}{type=\acronymtype, name={NCIA}, description={Network Camera Image Acquisition}, first = {Network Camera Image Acquisition (NCIA)}}
\newglossaryentry{api}{type=\acronymtype, name={API}, description={Application Programming Interface}, first = {Application Programming Interface (API)}}

\newglossaryentry{NDS}{type=\acronymtype, name={NDS}, description={Naturalistic Driving Study}, first={Naturalistic Driving Study (NDS)}, firstplural={Naturalistic Driving Studies (NDS)}}

\newglossaryentry{NDD}{type=\acronymtype, name={NDD}, description={Naturalistic Driving Data}, first={Naturalistic Driving Data (NDD)}, firstplural={Naturalistic Driving Data (NDD)}}

\newglossaryentry{NTD}{type=\acronymtype, name={NTD}, description={Naturalistic Trajectory Data}, first={Naturalistic Trajectory Data (NTD)}, firstplural={Naturalistic Trajectory Data (NTD)}}

\newglossaryentry{ecdf}{type=\acronymtype, name={ecdf}, description={empirical cumulative distribution function}, first = {empirical cumulative distribution function (ecdf)}}

\newglossaryentry{cdf}{type=\acronymtype, name={cdf}, description={cumulative distribution function}, first = {cumulative distribution function (cdf)}}

\newglossaryentry{pdf}{type=\acronymtype, name={pdf}, description={probability density function}, first = {probability density function (pdf)}}
\newglossaryentry{FOT}{type=\acronymtype, name={FOT}, description={Field Operations Test is a test method to assess the efficiency, quality, robustness and acceptance of e.g. \glsfirst{ADAS}}, first={Field Operations Test (FOT)}, firstplural={Field Operations Tests (FOT)}}

\newglossaryentry{SUT}{type=\acronymtype, name={SUT}, description={System under Test}, first={System under Test (SUT)}, firstplural={Systems under Test (SUT)}}

\newglossaryentry{OWL}{type=\acronymtype, name={OWL}, description={Web Ontology Language}, first={Web Ontology Language (OWL)}}

\newglossaryentry{RDF}{type=\acronymtype, name={RDF}, description={Resource Description Framework}, first={Resource Description Framework (RDF)}}

\newglossaryentry{SWRL}{type=\acronymtype, name={SWRL}, description={Semantic Web Rule Language}, first={Semantic Web Rule Language (SWRL)}}
\newglossaryentry{RWTD}{type=\acronymtype, name={RWTD}, description={Real World Test-Drive}, first={Real World Test-Drive (RWTD)}}
\newglossaryentry{KS}{type=\acronymtype, name={KS}, description={Knowledge-based System}, first={Knowledge-based System (KS)}, plural={Knowledge-based Systems (KS)}}

\newglossaryentry{KB}{type=\acronymtype, name={KB}, description={Knowledge Base}, first={Knowledge Base (KB)}, plural={Knowledg Base (KB)}}

\newglossaryentry{FSM}{type=\acronymtype, name={FSM}, description={Finite State Machine}, first={Finite State Machine (FSM)}, plural={Finite State Machines (FSM)}}

\newglossaryentry{DAS}{type=\acronymtype, name={DAS}, description={Driver Assistance System}, first={Driver Assistance System (DAS)}, plural={Driver Assistance Systems (DAS)}}

\newglossaryentry{TTC}{type=\acronymtype, name={TTC}, description={time to collision}, first={time to collision (TTC)}}

\newglossaryentry{TIV}{type=\acronymtype, name={TIV}, description={intervehicular time}, first={intervehicular time (TIV)}}
\newglossaryentry{CAV}{type=\acronymtype, name={CAV}, description={connected and automated vehicles}, first={connected and automated vehicles (CAV)}, firstplural={Connected and Automated Vehicles (CAVs)}}

\newglossaryentry{TTR}{type=\acronymtype, name={TTR}, description={time to react}, first={time to react (TTR)}}
\newglossaryentry{PET}{type=\acronymtype, name={PET}, description={Post
            Enroachment Time}, first={Post Enroachment Time (PET)}}
\newglossaryentry{GHG}{type=\acronymtype, name={GHG}, description={greenhouse gases}, first={greenhouse gases (GHG)}}
\newglossaryentry{ULDET}{type=\acronymtype, name={ULDET}, description={Unsupervised Lane change DeTection}, first={ULDeT (Unsupervised Lane change Detection)}}

\usepackage{transparent}
\usepackage{import}
\graphicspath{{figures/}}
\DeclareGraphicsExtensions{.pdf}

\newcommand{\etal}{{\it et~al. }}

\newcommand{\eg}{\textit{e.g.}}
\newcommand{\ie}{\textit{i.e.}}

\definecolor{dlrblue}{rgb}{0., 0.46, 0.73}
\definecolor{dlrgreen}{rgb}{0.51, 0.58, 0.35}
\definecolor{dlryellow}{rgb}{0.92, 0.72, 0.09}
\definecolor{dlrred}{rgb}{0.70, 0.25, 0.24}

\newcommand\figref[1]{Fig.~\ref{fig:#1}}
\newcommand\tabref[1]{TABLE~\ref{#1}}

\DeclareMathAlphabet{\mathcal}{OMS}{cmsy}{m}{n}

\newcolumntype{Y}{>{\hsize=0.25\hsize\arraybackslash} X}
\newcolumntype{C}{>{\centering\arraybackslash}X}

\hypersetup{hidelinks=true}

\title{\LARGE \bf Extraction and Analysis of Highway On-Ramp Merging Scenarios from Naturalistic Trajectory Data}

\author{Lars Klitzke\orcidlink{0000-0001-9362-707X}\authorrefmark{1}, Kay Gimm\orcidlink{0000-0002-5136-685X}\authorrefmark{1}, Carsten Koch\authorrefmark{2} and Frank
K\"oster\authorrefmark{3}
	\thanks{\authorrefmark{1}Lars Klitzke and Kay Gimm are with the
	German Aerospace Center (DLR), Institute of Transportation Systems,
	Braunschweig, Germany, {\tt \{lars.klitzke, kay.gimm\}@dlr.de}.}%
	\thanks{\authorrefmark{2}Carsten Koch is with the Hochschule Emden/Leer,
	University of Applied Sciences, Emden, Germany, {\tt
	carsten.koch@hs-emden-leer.de}.}%
	\thanks{\authorrefmark{3}Frank K\"oster is with the
	German Aerospace Center (DLR), Institute for AI Safety and Security, Sankt Augustin / Ulm, Germany, {\tt frank.koester@dlr.de}.}
}

\begin{document}
\maketitle
\thispagestyle{empty}
\pagestyle{empty}

\begin{abstract}
	\glsfirstplural{CAV} are envisioned to transform the future industrial and private transportation sectors. However, due to the system's enormous complexity, functional verification and validation of safety aspects are essential before the technology merges into the public domain. Therefore, in recent years, a scenario-driven approach has gained acceptance, emphasizing the requirement of a solid data basis of scenarios.

The large-scale research facility \glsfirst{TFNDS} enables the provision of ample information for a database of scenarios on highways. For that purpose, however, the scenarios of interest must be identified and extracted from the collected \gls{NTD}. This work addresses this problem and proposes a methodology for on-ramp scenario extraction, enabling scenario categorization and assessment. An \gls{HMM} and \gls{DTW} is utilized for extraction and a decision tree with the \gls{SMoS} \gls{PET} for categorization and assessment. The efficacy of the approach is shown with a dataset of \gls{NTD} collected on the \glsname{TFNDS}.
\end{abstract}


\section{Introduction}
\label{sec:introduction}
The transport sector has a significant impact on the environment. In 2017, the mobility sector emitted $18.5\%$ of the total \gls{GHG} in Germany \cite{gunther2019resource}. Especially with the Paris agreement in 2016, the automotive industry faces a massive challenge by reducing the \gls{GHG} of their fleets to become carbon neutral. Although a general rethinking of mobility is needed, several emerging technologies can help achieve this goal.

In recent years, researchers focussed on the topic of \gls{CAV} due to the availability of the required sensors for environment perception and infrastructure technology for communication. The communication between vehicles (V2V) or vehicles and infrastructure (V2I) enables \gls{CAV} to conduct their driving task more efficiently and thus reducing the impact on the environment \cite{KOPELIAS2020135237}. Moreover, \glsplural{CAV} shall provide a safer \cite{PAPADOULIS201912}, more energy-efficient \cite{7548168}, and comfortable driving experience.

However, before \glsplural{CAV} merge into the market, a homologation is required. This is addressed by several research projects such as PEGASUS, ENABLE-3S and their successors VVM and SET Level\footnote{www.pegasusprojekt.de/en, www.enable-s3.eu, www.vvm-projekt.de/en, https://setlevel.de/en}. The systems' functional correctness  needs to be verified, which is hard to accomplish with real-world drivings only \cite{WinnerPegasus2018}. Hence, scenario-driven simulation-based testing is applied extensively throughout the development cycle \cite{2018Hallerbach, 9330510}.

A scenario of high interest in the context of \gls{CAV} on highways is the on-ramp merging scenario with a vehicle merging onto the mainline from an on-ramp lane. Merging vehicles can affect the overall traffic flow, and those situations can become critical in case of, \eg{}, a short on-ramp lane, high occupancy of the mainline, occluded field-of-view, or high relative velocities. Furthermore, vehicles participating in this scenario often need to collaborate to resolve the situation, making this scenario particularly interesting in mixed traffic situations of \gls{CAV} and traditional vehicles. However, an extensive catalog of scenarios is essential for such analysis \cite{damm2018formal}, with \glsfirst{NTD} from the real-world providing valuable information.

\begin{figure}[h]
    \centering
    \includegraphics[width=\linewidth]{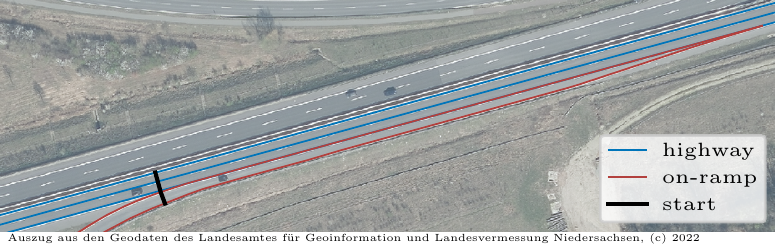}
    \caption{The on-ramp section near Cremlingen of the \gls{TFNDS}. There are two mainline lanes denoted as {\color{dlrblue} highway} and one acceleration lane denoted as {\color{dlrred} on-ramp}.}
    \label{fig:aggregated_map}
\end{figure}

The \glsfirst{DLR} constructed a large-scale research facility named \gls{TFNDS} to test technology in the context of connected and automated driving (see \figref{aggregated_map} for the on-ramp section near Cremlingen). The camera-based infrastructure for traffic analysis on the  A39 highway with a length of $7 km$ allows tracking objects on the highway and providing information about a \gls{VuT} such as the pose, dimension, vehicle type and its environment in specific scenarios. For scenario-based analysis of a \gls{VuT} and situation assessment, \eg{}, in terms of criticality using \glsfirstplural{SMoS} such as the \gls{TTC} \cite{9330510} or \gls{PET} \cite{9069432}, however, the extraction of these scenarios, also referred to as scenario mining \cite{elrofai2018streetwise}, is mandatory.

\begin{figure*}[b]
    \centering
    \includegraphics[width=\linewidth]{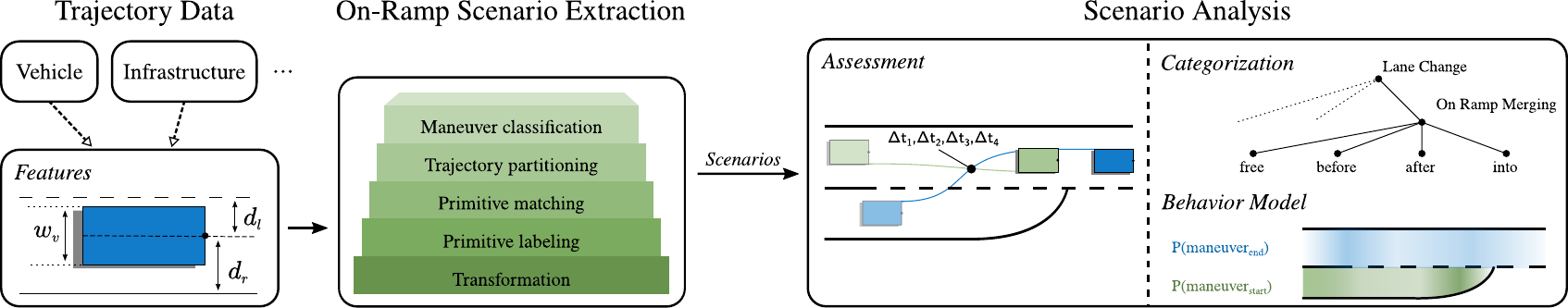}
    \caption{The proposed framework for on-ramp scenario extraction and
        analysis. The trajectory data can be supplied by various sources and just
        needs to be converted to provide the required \textit{Features} which are
        used for scenario extraction. The scenarios are assessed, categorized and the on-ramp behavior is characterized in the
        \textit{Scenario Analysis} module.}
    \label{fig:new_framework}
\end{figure*}

\subsection{Related work}
The extraction of scenarios has been a broadly discussed topic in the research community for several years and has become even more relevant with the recent shift to a scenario-driven validation approach. A short overview of recent activities in this domain is given in the following.

Various approaches exist in literature to tackle the problem of scenario extraction in trajectory data. In general, the aim is to find certain characteristic time intervals in a multivariate time series. Supervised learning methods such as the \gls{SVM} and \gls{ANN} were proposed by \cite{Kumar_SVM_2013, SVm_RBF, Ramyar_OneClassSVM, 2018MonotANN}. However, the inherent drawback of these approaches is the training of (hyper) parameters based on ground truth data.

A comparative study for scenario extraction is conducted by Erdogan \etal{} \cite{2019Erdogan} with a rule-based, supervised (\gls{RNN}) and unsupervised learning ($k$-NN and \gls{DTW}) method. Although the \gls{RNN} is superior to the unsupervised
learning method, it is more prone to sensor noise \cite{2019Erdogan} which is
inevitable in real-world data.

An alternative to supervised learning-based methods are approaches for clustering, \ie, unsupervised learning methods. Montanari \etal{} \cite{9304560} slice time-series into chunks based on artificially created signals with empirically defined parameters and \gls{AHC}. The final interpretation of clusters is performed manually to find different variants of lane changes. This problem is inherent with non-deterministic unsupervised learning methods such as \gls{AHC}, Random  Forest Clustering \cite{8569682} or \gls{HMM} \cite{klitzke_itsc2020} and post-processing steps are required to recover the semantics either manually or automatically \cite{8506402}. For instance, Elspas \etal{} \cite{9272025} employ regular expressions to find maneuvers in the trajectory data after clustering it into symbolic states. However, they show that crafting the patterns has a significant impact on identification accuracy.

\subsection{Contribution}

This work proposes a novel approach for extraction and analysis of on-ramp merging scenario from \gls{NTD} contributing to setting up an extensive dataset of on-ramp merging scenarios for testing. Therefore, this work follows the divide and conquer strategy by decomposing an on-ramp maneuver into substates, the \textit{primitives}. This abstract representation allows finding and classifying merging maneuvers robustly, as shown in a previous work \cite{klitzke_itsc2020} for lane-change maneuvers. An \gls{HMM} is employed to represent the trajectory in the primitive domain, and \gls{DTW} is used to compare the partitioned trajectory with patterns representing different variants of an on-ramp maneuver in the primitive domain.

The approach is applied to extract on-ramp scenarios from \gls{NTD} of the \gls{TFNDS} and analyze the on-ramp behavior of traffic participants. These insights are valuable especially in the context of mixed-traffic of \gls{CAV} and traditional vehicles. Therefore, the \gls{SMoS} \gls{PET} is used for finding critical situations and to create a decision tree for categorizing merging scenarios according to the occupancy of the mainline.

\subsection{Paper structure}
The remaining paper is structured as follows. In Section~\ref{sec:framework}, the proposed approach for on-ramp scenario extraction is presented. In Section~\ref{sec:categorization_and_assessment}, the scenario analysis module is introduced, which allows categorizing and assessing the extracted scenarios. An evaluation of the scenario extraction method and an analysis of on-ramp merging scenarios follows in Section~\ref{sec:evaluation}. Finally, the work concludes with a summary and outlook on future work in Section~\ref{sec:conclusion}.

\section{On-Ramp Merging Scenario Extraction}
\label{sec:framework}
For the extraction and analysis of on-ramp merges in \gls{NTD} provided by the \gls{TFNDS}, this work proposes a methodology depicted in \figref{new_framework}. In the following, the fundamental assumptions on on-ramp maneuvers are discussed, and the approach for on-ramp scenario extraction is presented.

\subsection{On-ramp merging maneuver}
\label{subsec:merge-maneuver}
An on-ramp merging maneuver is a lane-change maneuver but in a different context. In fact, a road user performing an on-ramp maneuver is, in this work, assumed to drive on the acceleration lane of a highway's on-ramp section and performs a lane-change to the left lane (see \figref{aggregated_map}). A lane-change is in this work partitioned into multiple states, the \textit{driving primitives}. This allows breaking down the complexity of a lane-change maneuver into more trivial driving actions which can be identified more easily. In addition, they are the building blocks for maneuver modelling and allow analyzing the behavior of traffic participants in the maneuver's primitive domain space. \cite{klitzke_itsc2020}

\begin{figure}[htb]
	\centering
	\def\svgwidth{1\linewidth}
	\fontsize{8pt}{0pt}
\begingroup%
  \makeatletter%
  \providecommand\color[2][]{%
    \errmessage{(Inkscape) Color is used for the text in Inkscape, but the package 'color.sty' is not loaded}%
    \renewcommand\color[2][]{}%
  }%
  \providecommand\transparent[1]{%
    \errmessage{(Inkscape) Transparency is used (non-zero) for the text in Inkscape, but the package 'transparent.sty' is not loaded}%
    \renewcommand\transparent[1]{}%
  }%
  \providecommand\rotatebox[2]{#2}%
  \newcommand*\fsize{\dimexpr\f@size pt\relax}%
  \newcommand*\lineheight[1]{\fontsize{\fsize}{#1\fsize}\selectfont}%
  \ifx\svgwidth\undefined%
    \setlength{\unitlength}{236.24997837bp}%
    \ifx\svgscale\undefined%
      \relax%
    \else%
      \setlength{\unitlength}{\unitlength * \real{\svgscale}}%
    \fi%
  \else%
    \setlength{\unitlength}{\svgwidth}%
  \fi%
  \global\let\svgwidth\undefined%
  \global\let\svgscale\undefined%
  \makeatother%
  \begin{picture}(1,0.32658103)%
    \lineheight{1}%
    \setlength\tabcolsep{0pt}%
    \put(0,0){\includegraphics[width=\unitlength,page=1]{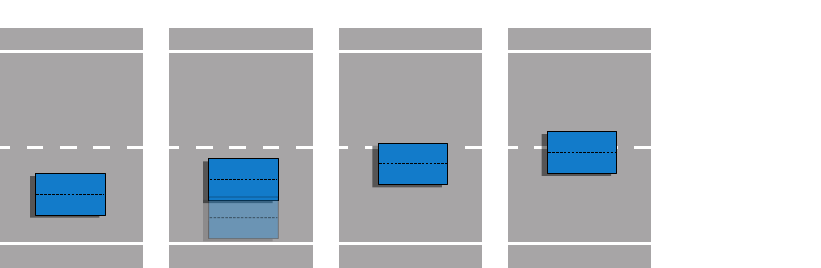}}%
    \put(0.08733394,0.30306912){\color[rgb]{0,0,0}\makebox(0,0)[t]{\lineheight{0}\smash{\begin{tabular}[t]{c}Idle\end{tabular}}}}%
    \put(0,0){\includegraphics[width=\unitlength,page=2]{states.pdf}}%
    \put(0.29368316,0.30306912){\color[rgb]{0,0,0}\makebox(0,0)[t]{\lineheight{0}\smash{\begin{tabular}[t]{c}Approach\end{tabular}}}}%
    \put(0.50003238,0.30330618){\color[rgb]{0,0,0}\makebox(0,0)[t]{\lineheight{0}\smash{\begin{tabular}[t]{c}Cross\end{tabular}}}}%
    \put(0.7063816,0.30306912){\color[rgb]{0,0,0}\makebox(0,0)[t]{\lineheight{0}\smash{\begin{tabular}[t]{c}Change\end{tabular}}}}%
    \put(0.91273226,0.30378026){\color[rgb]{0,0,0}\transparent{0.40000001}\makebox(0,0)[t]{\lineheight{0}\smash{\begin{tabular}[t]{c}Depart\end{tabular}}}}%
    \put(0,0){\includegraphics[width=\unitlength,page=3]{states.pdf}}%
  \end{picture}%
\endgroup%

	\caption{A lane-change is represented with four driving primitives \cite{klitzke_itsc2020}. From a lane-oriented point of view the state textit{Depart} is equal to \textit{Change} but in the opposite direction.}
	\label{fig:states}
\end{figure}

From a lane-oriented point of view, the maneuver is divided into four distinct states (see \figref{states}). The vehicle is in the state \textit{Idle} if it is in the middle of the lane. Moving towards the left lane marking, the vehicle enters the state \textit{Approach} until it crosses it, where it changes to state \textit{Cross}. The state \textit{Change} denotes that the vehicle transitions to the target as long as the majority of the vehicle is on the source lane. Then, instead of changing to the state \textit{Depart}, it will remain in the state \textit{Change} but is associated to the target lane. Thus, the domain $D$ of driving primitives is $D = \{\text{\textit{Idle}}, \text{\textit{Approach}}, \text{\textit{Cross}}, \text{\textit{Change}}\}$ for the lane-change or on-ramp maneuver. The aim is to, at first, transform the trajectory into the domain of driving primitives. By doing so, we can define different variants of a maneuver on the more abstract driving primitive domain level.

Let us assume that the trajectory of a road user is transformed into the feature space depicted in \figref{new_framework}. Hence, let $X = \mathbf{x}_1, \mathbf{x}_2, \dots, \mathbf{x}_n$ be the trajectory in the feature space so that $\mathbf{x} = [d_l, d_r, w]^T$ denotes the distance to the next left $d_l$ and next right $d_r$ marking, and $w$ the vehicle's width. The aim is to find the most likely sequence $\mathcal{D} = \mathcal{D}_1, \mathcal{D}_2, \dots, \mathcal{D}_n$ with $\mathcal{D}$ as the primitive's index in domain $D$, that represents the trajectory in the primitive domain defined by
\begin{equation}
	\arg \max_{\mathcal{D}}\, P( \mathcal{D} \mid X)
	\label{equ:hmm-problem}
\end{equation}
with $P(\mathcal{D} \mid X)$ denoting the probability of \textit{observing} the series of primitives $\mathcal{D}$ given a trajectory $X$. For solving \eqref{equ:hmm-problem}, we use an \gls{HMM} since it allows inferring a system's hidden state based on its emitted information. In this particular case, the hidden states are the primitives and the information that the system emits is the trajectory. A \gls{HMM} is defined as $\lambda = (\bm{A}, \bm{B}, \bm{\pi})$ with $\bm{A}$ as the state transition probabilities, $\bm{B}$ the matrix of observation probability conditioned on the current hidden state and $\bm{\pi}$ the initial state probabilities \cite{18626}. Since the observations are non-discrete we model each state's emissions with a \gls{GMM}. Assuming that $\lambda$ is a trained \gls{HMM} with Gaussian emissions, the Viterbi algorithms can be utilized for solving the problem defined in \eqref{equ:hmm-problem} \cite{bilmes1998gentle}. That is, finding the optimal sequence of primitives given the observations.

Note that the features depicted in \figref{new_framework} are not used directly, but a transformation is applied that allows supporting different vehicles and lanes. The features are the distance from the vehicle's center to the lane's center $d_c$ and a marker $\kappa$ indicating the vehicle is crossing a lane border. The \gls{HMM} parameters used in this work are shown in \tabref{tab:dp_hmm_example} taken from a previous work \cite{klitzke_itsc2020}. The state transition matrix is denoted as \textit{Transition probability} and the state emission via \textit{State distribution}. Note that only the two adjacent primitives for each primitive have a significant transition likelihood which is coherent with our expectations and the states depicted in \figref{states} since a vehicle should not switch between, \eg, the primitives \textit{Approach} and \textit{Change} and thus skipping \textit{Cross}.

\begin{table}[hb]
	\centering
	\caption{The \gls{HMM} parameter used for finding the most likely sequence of lane-change primitives given a trajectory.}

	\label{tab:dp_hmm_example}
	\begin{tabularx}{\linewidth}{@{}lCCCCcC@{}}
		\toprule
		{}        & \multicolumn{4}{c}{Transition probability $\bm{A}$} & \multicolumn{2}{c}{State distribution  $\bm{B}$}                                                        \\
		Primitive & Idle                                                & Approach                                         & Cross     & Change    & $d_c$             & $\kappa$ \\
		\midrule
		Idle      & \(98.94\)                                           & \(1.03\)                                         & \(0.03\)  & \(0.00\)  & \(0.09 \pm 0.06\) & \(0.00\) \\
		Approach  & \(1.46\)                                            & \(97.53\)                                        & \(1.01\)  & \(0.00\)  & \(0.33 \pm 0.08\) & \(0.00\) \\
		Cross     & \(0.47\)                                            & \(8.28\)                                         & \(86.17\) & \(5.08\)  & \(0.53 \pm 0.09\) & \(1.00\) \\
		Change    & \(0.00\)                                            & \(0.33\)                                         & \(5.98\)  & \(93.69\) & \(0.89 \pm 0.11\) & \(1.00\) \\
		\bottomrule
	\end{tabularx}
\end{table}

\subsection{Scenario extraction}
A road user on the acceleration lane might change arbitrarily between the driving primitive states, and there could be multiple interesting situations to study. For instance, a driver on the acceleration lane could misjudge the velocity of a vehicle on the mainline and starts the merging maneuver. The driver reevaluates the situation during the maneuver and aborts it by changing back to the acceleration lane to prevent a critical situation. Consequently, not only the on-ramp maneuver is interesting, but also the canceled one. Due to that, we need to find the sequences within the trajectory representing these situations. If $\mathcal{D} = \mathcal{D}_1, \mathcal{D}_2, \dots, \mathcal{D}_n$ is the series of primitives, it will be divided into sequences $\mathcal{S} = \{\mathcal{S}_1, \mathcal{S}_2, \dots, \mathcal{S}_k\}$ with $k \ll n$ such that  $\forall \mathcal{S}_i \in \mathcal{S} : \mathcal{S}_i^k \ge \xi$ with $\mathcal{S}_i^k$ as the $k^{th}$ primitive of the $i^{th}$ partition and  $\xi$ defining the lower primitive bound. For instance, if $\xi =2$ a sequence represents a situation where a road user is either in the state \textit{Cross} or changing between \textit{Cross} and \textit{Change}. In fact, the work uses $\xi=2$ for on-ramp maneuver extraction since the aim is to analyze the merging maneuver focussing on the interval where the vehicle transitions between the lanes.

For scenario identification, the sequences $\mathcal{S} = \mathcal{S}_1, \mathcal{S}_2, \dots$ need to be classified according to the relevant on-ramp variants. If $\mathcal{H} = \mathcal{H}_0, \mathcal{H}_1, \dots$ is the set of patterns representing the variants in the primitive domain, the aim is to find the most likely pattern $\widehat{\mathcal{H}}$ defined by

\begin{equation}
	\widehat{\mathcal{H}} = \arg \max_{\mathcal{H}}\; d\left(\mathcal{S}_i, \mathcal{H}\right)
	\label{equ:most-likely-sequence}
\end{equation}

with $d(\mathcal{S}_i, \mathcal{H}_j)$ denoting the \textit{similarity} between both series. For time-series similarity estimation in \eqref{equ:most-likely-sequence}, \gls{DTW} is employed. For further details, the interested reader is referred to \cite{klitzke_itsc2020}.

\section{Scenario categorization and assessment}
\label{sec:categorization_and_assessment}
The identified on-ramp merging scenarios using the approach presented in the previous section are processed in the \textit{Scenario Analysis} component for situation assessment, scenario categorization and behavior modeling as depicted in \figref{new_framework}. For the assessment of scenarios in terms of the criticality so called \glsfirstplural{SMoS} are typically utilized. For lane changes, the \gls{PET} is broadly used \cite{behbahani2015framework, 9069432} for that purpose and thus chosen for this work since it allows assessing a certain situation but also provides information for further scenario categorization. In fact, this work will use a decision tree for scenario categorization based on the mainline's occupancy and the \gls{PET}.

\subsection{Situation assessment}
The \gls{PET} is the time difference between the first object leaving and the second object entering a conflict area \cite{2003douglas}. In the following, the first is denoted as the \textit{ego vehicle} and the latter as the \textit{challengers} as defined in \cite{2019weber}. The lower the \gls{PET} between two objects, the higher the potential criticality. This work assumes that the conflict area is defined by the four intersection points of the ego vehicle's rear left and front left edges to the rear right and front ride edges of the challengers (see \figref{new_framework}). Let $\mathcal{C} = \mathcal{C}_1, \mathcal{C}_2, \dots$ be the challengers on the mainline with $\mathcal{C}_i = \mathbf{c}_1, \mathbf{c}_2, \dots$ as the challenger's trajectory that consists of the vehicle's front right and rear right position. 
Furthermore, let $\mathcal{X} = \mathbf{x}_1, \mathbf{x}_2, \dots$ be the ego vehicle's trajectory with its front left and rear left position. 
Then, $\mathcal{P} = \{\mathbf{p}_1, \mathbf{p}_2, \mathbf{p}_3, \mathbf{p_4}\}$ are the four intersection points between a challengers and the ego vehicle's trajectory.
For \gls{PET} estimation, the difference in time of arrival $T_{\mathcal{C}_i} = \{\Delta t_1, \Delta t_2, \Delta t_3, \Delta t_4\}$ with $\Delta t = t_{ego} - t_{\mathcal{C}_i}$ and $t_{ego}$ of the ego vehicle's and $t_{\mathcal{C}_i}$ the challenger's arrival time at the intersection point $\mathbf{p} \in \mathcal{P}$ is calculated. Given the time-differences, the $\operatorname{PET}(\mathcal{C}_i)$ between the ego vehicle and the candidate $\mathcal{C}_i$ is defined as
\begin{equation}
	\begin{aligned}
		\operatorname{PET}(\mathcal{C}_i) =\{ \Delta t \mid \,  \Delta t \in T_{\mathcal{C}_i} \wedge \lvert \Delta t \rvert = \min\lvert T_{\mathcal{C}_i}\rvert \}
	\end{aligned}
	\label{equ:pet}
\end{equation}
with $\min\lvert T_{\mathcal{C}_i}\rvert$ the minimum absolute arrival time difference. The \gls{SMoS} \gls{PET} of on-ramp scenarios can be employed to find critical scenarios as exemplarily shown in Section~\ref{sec:evaluation}.

\subsection{Scenario categorization}
\label{subsec:categorization}
\begin{figure}[htb]
	\centering
	\def\svgwidth{1\linewidth}
	\fontsize{8pt}{0pt}
	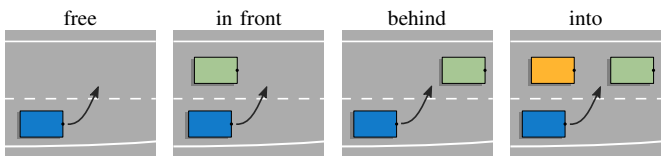
	\caption{On-ramp merging scenarios are specialized into four classes
	according to the position of challengers}
	\label{fig:lane_change_types}
\end{figure}

An on-ramp scenario can be further categorized according to the challengers \cite{9069432}. In this work, only the challengers on the mainline are considered for categorization that are in the vicinity of the ego vehicle at that point in time where the ego vehicle performs a merging maneuver and is entering the primitive \texttt{Cross}. In particular, an on-ramp scenario is in this work specialized into four subtypes: \textit{free}, \textit{in front}, \textit{behind} and \textit{into} as depicted in \figref{lane_change_types}.

For this categorization, a decision tree is created based on the presence of challengers on the mainline and the \gls{PET} between the ego vehicle and all challengers of a scenario. The \gls{PET} is used since the sign of the \gls{PET} denotes the order of entering and leaving the conflict zone. The PET defined in \eqref{equ:pet} is positive if the ego vehicles arrives the intersection point after the challenger and negative vice versa. Hence, a scenario is classified as \textit{behind} or \textit{in front} if there is only one challenger $|\mathcal{C}| = 1$ and $\operatorname{PET}(\mathcal{C}_1) > 0$ or $\operatorname{PET}(\mathcal{C}_1) < 0$ respectively. If there are at least two challengers $|\mathcal{C}| > 2$ and $\exists \mathcal{C}_i \in \mathcal{C}: \operatorname{PET}(\mathcal{C}_i) > 0 \wedge \mathcal{C}_i \in \mathcal{C}: \operatorname{PET}(\mathcal{C}_i) < 0$ the lane-change is classified as \textit{into}. If there is no challenger on the mainline, the lane-change is classified as \textit{free}. Note that this work assumes that there is no collision between challengers and the ego vehicle.

\subsection{Driver behavior model}
\label{subsec:modelling}
The identification of on-ramp merging scenarios enables to derive behavioral models of the drivers, which can be employed to generate trajectories for more extensive simulation-based testing, \eg, in Carla \cite{2017_carla} or SUMO \cite{dlr71460}. This is especially of interest in the context of mixed traffic of \gls{CAV} and traditional vehicles since \gls{CAV} have to predict the behavior in this situations.

\begin{figure}[b]
	\centering
	\def\svgwidth{0.9\linewidth}
	\fontsize{8pt}{0pt}
\begingroup%
  \makeatletter%
  \providecommand\color[2][]{%
    \errmessage{(Inkscape) Color is used for the text in Inkscape, but the package 'color.sty' is not loaded}%
    \renewcommand\color[2][]{}%
  }%
  \providecommand\transparent[1]{%
    \errmessage{(Inkscape) Transparency is used (non-zero) for the text in Inkscape, but the package 'transparent.sty' is not loaded}%
    \renewcommand\transparent[1]{}%
  }%
  \providecommand\rotatebox[2]{#2}%
  \newcommand*\fsize{\dimexpr\f@size pt\relax}%
  \newcommand*\lineheight[1]{\fontsize{\fsize}{#1\fsize}\selectfont}%
  \ifx\svgwidth\undefined%
    \setlength{\unitlength}{288bp}%
    \ifx\svgscale\undefined%
      \relax%
    \else%
      \setlength{\unitlength}{\unitlength * \real{\svgscale}}%
    \fi%
  \else%
    \setlength{\unitlength}{\svgwidth}%
  \fi%
  \global\let\svgwidth\undefined%
  \global\let\svgscale\undefined%
  \makeatother%
  \begin{picture}(1,0.375)%
    \lineheight{1}%
    \setlength\tabcolsep{0pt}%
    \put(0,0){\includegraphics[width=\unitlength,page=1]{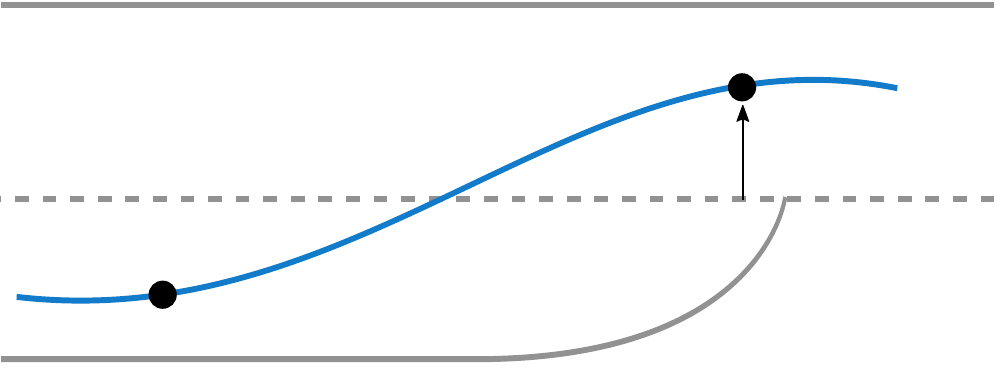}}%
    \put(0.75443189,0.22086699){\color[rgb]{0,0,0}\makebox(0,0)[lt]{\lineheight{0}\smash{\begin{tabular}[t]{l}$g(x_{t_2}, r)$\end{tabular}}}}%
    \put(0.71365668,0.20674014){\color[rgb]{0,0,0}\makebox(0,0)[rt]{\lineheight{0}\smash{\begin{tabular}[t]{r}$f(x_{t_2}, r)$\end{tabular}}}}%
    \put(0.17104373,0.13324805){\color[rgb]{0,0,0}\makebox(0,0)[lt]{\lineheight{0}\smash{\begin{tabular}[t]{l}$g(x_{t_1}, r)$\end{tabular}}}}%
    \put(0,0){\includegraphics[width=\unitlength,page=2]{frenet_space.pdf}}%
    \put(0.14756253,0.1290268){\color[rgb]{0,0,0}\makebox(0,0)[rt]{\lineheight{0}\smash{\begin{tabular}[t]{r}$f(x_{t_1}, r)$\end{tabular}}}}%
  \end{picture}%
\endgroup%

	\caption{The trajectory in Frenét reference frame defined by the left border $r$ of the acceleration lane.}
	\label{fig:frenet_frame}
\end{figure}

In this work, the focus is on the distribution of the merging maneuver's \textit{start} and \textit{end} position as a simple use-case to demonstrate the potential of collecting scenarios with the proposed approach. For that purpose, let the merging-maneuver $m = (p_1, p_2)$ be defined by its start $p_1$ and end position $p_2$. However, instead of using the locations in a Cartesian Coordinate system, the positions are converted into the Frenét reference frame defined by the left lane border of the acceleration lane depicted in \figref{frenet_frame}.
\begin{figure*}[!b]
	\includegraphics[width=0.33\textwidth]{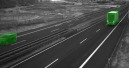}\hfill
	\includegraphics[width=0.33\textwidth]{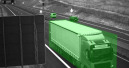}\hfill
	\includegraphics[width=0.33\textwidth]{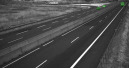}\\
	\includegraphics[width=0.33\textwidth]{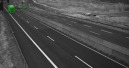}\hfill
	\includegraphics[width=0.33\textwidth]{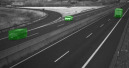}\hfill
	\includegraphics[width=0.33\textwidth]{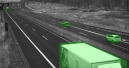}
	\caption{Images of the on-ramp section near Cremlingen of the \gls{TFNDS} with overlayed bounding boxes. The dataset used in this work is collected in that area for scenario extraction and analysis.}
	\label{fig:cremlingen}
\end{figure*}
This is an established method for, e.g., trajectory planning since it allows defining a trajectory relative to the road \cite{5509799}. That is, the location $\mathbf{x}_t$ at time $t$ represented in a Cartesian Coordinate system is converted into a dynamic reference frame specified by the lane boundary's representation $r$ according to \cite{5509799} by
\begin{equation}
	\mathbf{x}_t = f\left(\mathbf{x}_t, r\right) + g\left(\mathbf{x}_t, r\right)
	\label{equ:frenet-space}
\end{equation}
with $f\left(\mathbf{x}_t, r\right)$ denoting the distance of the position $\mathbf{x}_t$ on the lane $r$ and  $g\left(\mathbf{x}_t, r\right)$ the perpendicular distance between $\mathbf{x}_t$ and its projected sibling on the lane. Since only the relative position on the lane is relevant to describe the maneuver start and end position, the second component of \eqref{equ:frenet-space} is neglected. Furthermore, since merging maneuvers can only occur when the acceleration lane meets the mainline, the start of the lane is defined at the point where the acceleration lane's left lane border joins with the rightmost border of the highway lanes as denoted as the \textit{start} line in \figref{aggregated_map}.
Hence, the maneuver start and end position is given by
\begin{equation}
	m =\left(\frac{1}{\lvert r \rvert}f(\mathbf{x}_{t_1}, r) - \nu, \frac{1}{\lvert r \rvert}f(\mathbf{x}_{t_2}, r)- \nu\right)
	\label{equ:maneuer-position}
\end{equation}

with $\mathbf{x}_{t_1}, \mathbf{x}_{t_2}$ as the road user's location at the maneuver start and end respectively, $\nu$ the start position on the lane and $\lvert r \rvert$ the lane's length from the start point. The latter is employed for position normalization to ensure that the results are independent of the acceleration lane length. The maneuver's start and end positions defined in \eqref{equ:maneuer-position} is used in Section~\ref{sec:evaluation} for further analysis.

\section{Experiments}
\label{sec:evaluation}
The following section evaluates the proposed approach for on-ramp merging extraction with a dataset of the \gls{TFNDS}. The extracted scenarios are the basis for further analysis.

\subsection{Dataset}
The trajectory data used for the evaluation and analysis is collected at the on-ramp region at Cremlingen near Braunschweig in Germany, which is part of the \gls{TFNDS} (see \figref{aggregated_map}). The region is covered by four camera masts equipped with two stereo cameras. In \figref{cremlingen} a sample image per camera is shown; row-wise from the top left to the bottom right. Only the stereo camera pair pointing to the on-ramp is shown for the first and last camera mast. The trajectory data was collected on three different days and various daytimes. Note that the used sampling strategy was to collect snippets of $90$ seconds each $300$ seconds. Due to this, only trajectories are considered in the analysis having a length greater than half of the on-ramp lane's length to remove clipped trajectories.

\begin{figure}[h]
    \centering
    \includegraphics[width=\columnwidth]{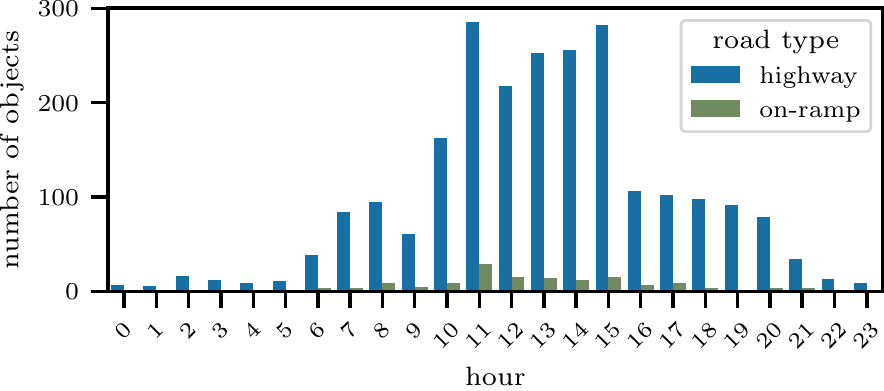}
    \caption{Distribution of the observed objects per road type on a hourly basis in the used dataset.}
    \label{fig:distribution}
\end{figure}

The road association of a traffic participant was estimated for qualitative analysis of the proposed extraction method. Therefore, it is assumed that a trajectory performs a merging maneuver if the trajectory starts on the acceleration but ends on any highway lane. In all other cases, the traffic participant is associated with the highway. After filtering, the total number of trajectories in the dataset is $2465$ with $2321$ objects associated with the highway lanes and $144$ the on-ramp lane (performing a merging maneuver). The distribution of the observed objects on an hourly basis is depicted in \figref{distribution}.

\subsection{Scenario extraction evaluation}
To evaluate the scenario extraction accuracy, the dataset is processed with the presented approach. Therefore, only objects driving on the on-ramp are considered.  As a result, the approach correctly extracts $137$ of the $144$ merging scenarios and thus achieves an accuracy of $95.14\%$.

\begin{figure}[h]
    \centering
    \includegraphics[width=\columnwidth]{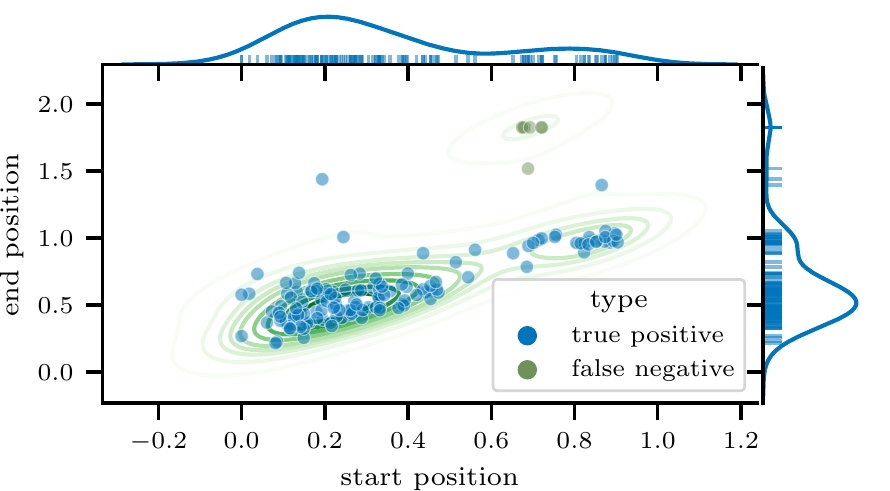}
    \caption{Overview of the correctly ($\bullet$) and incorrectly ($\star$)
        identified on-ramp merges start and end positions with object classes
        (color-encoded).}
    \label{fig:invalid_overview}
\end{figure}

It turns out that the approach fails in lane-change identification if a lane-change starts at the end of the on-ramp lane so that the object crosses both lane marking. This is depicted in \figref{invalid_overview} showing the relation between the start and end positions of all maneuvers. 
Note that all positions in \figref{invalid_overview} are relative to the start of the lane as defined in Section~\ref{subsec:modelling}.

It is evident from \figref{invalid_overview} that the difference between correctly and incorrectly identified maneuvers is the end position. That is, maneuvers starting at the end of the on-ramp lane with a start position greater than $0.6$ and ending beyond it with an end position greater $1.4$ are not identified.

Overall, although the approach failed to extract all scenarios, the performance is appropriate for the use-case at hand since no false positives exist. This is because the \gls{TFNDS} is collecting a vast amount of data, and missing scenarios is not critical but extracting misclassified scenarios is.  The latter is due to the required resources for scenario extracting and saving. Moreover, those misclassified scenarios need to be identified and removed before further analysis. Otherwise, wrong conclusions about the on-ramp merging behavior could be drawn.

\subsection{Merging behavior analysis}
\label{subsec:ramp_behavior}
\begin{figure}[h]
    \vspace{.5em}
    \includegraphics[width=1.0\columnwidth]{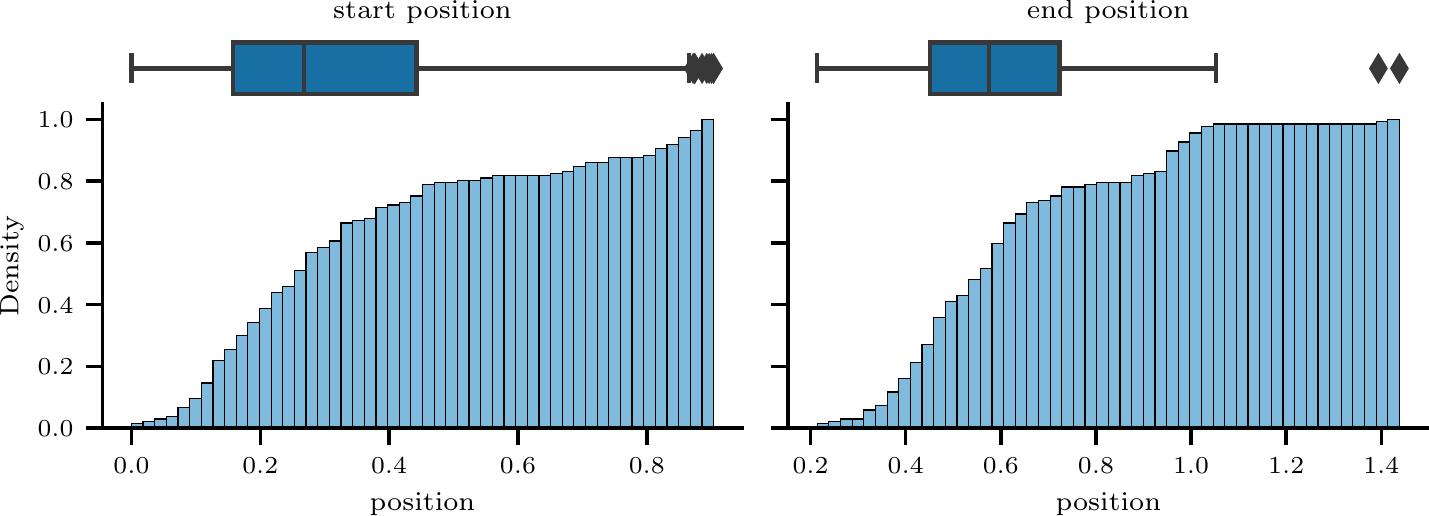}
    \caption{The \gls{cdf} of maneuver start and end positions relative to the start and normalized w.r.t the length of the acceleration lane. Only the correctly extracted maneuvers are considered.}
    \label{fig:ecdf_maneuver_start}
\end{figure}

The representation of on-ramp maneuvers with the starting and end position allows analyzing the merging behavior. As depicted in \figref{invalid_overview}, the distribution of start positions is bimodal with the majority of maneuver starting at the first half of the acceleration lane. This is also clearly visible in \figref{ecdf_maneuver_start} depicting the start and end position's \gls{cdf} of all correctly extracted maneuvers. That is, $25\%$ of all maneuvers have a start position smaller than $0.16$ and $50\%$ smaller than $0.27$. Hence, most maneuvers start within the first half of the acceleration lane. In fact, $79.56\%$ of all maneuvers have a start position smaller than $0.5$ and $86.13\%$ start within the first three quarters. The opposite is the case for the maneuver end positions with $1.46\%$ within the first quarter and $39.42\%$ in the first half. The most traffic participants finish the on-ramp maneuver within the second half ($53.28\%$) and almost all ($92.70\%$) within the course of the acceleration lane. Hence, $7.30\%$ are still conducting the maneuver although the acceleration lane has already merged.

\subsection{Scenario assessment}
The \texttt{Assessment} module depicted in \figref{new_framework} uses the \gls{PET} to relate the ego-vehicle on the on-ramp lane to challengers on the target lane. This enables to, for instance, find critical scenarios, \ie{}, scenarios with a low \gls{PET} between the ego-vehicle and a challenger.

\begin{figure}[htb]
    \centering
    \includegraphics[width=0.95\columnwidth]{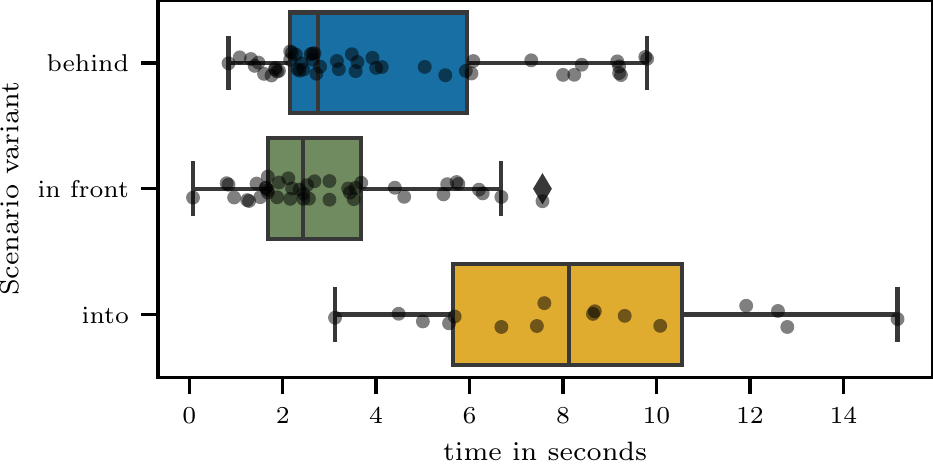}
    \caption{Distribution of the PET time for all identified on-ramp scenarios
        of type \textit{behind} and \textit{in front} and the accepted gap time for
        \textit{into}.}
    \label{fig:criticality_analysis}
\end{figure}

The \gls{PET} distribution for the scenario categories \textit{behind}, \textit{in front} and \textit{into} is depicted in \figref{criticality_analysis}. Note that the category \textit{free} is missing since there is no challenger on the mainline. For the category \textit{into} the sum of PETs values is used to represent each scenario with a single value which is related to the \gls{SMoS} accepted gap time.

\begin{figure}[h]
    \centering
    \includegraphics[width=\columnwidth]{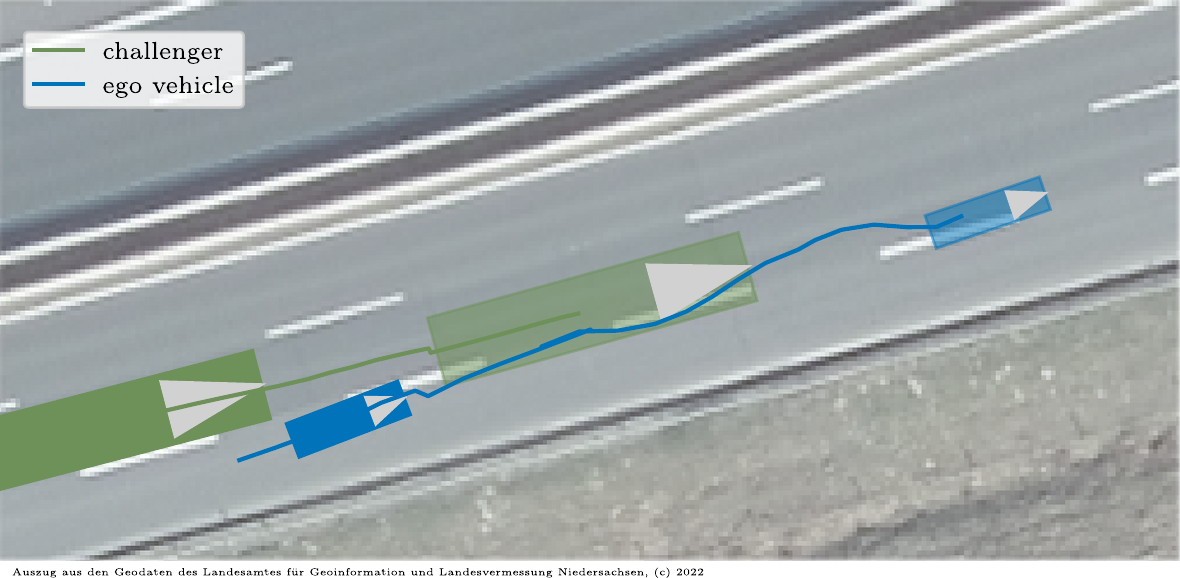}
    \caption{The identified critical on-ramp scenario at the start of the
        lane change maneuver and one second later (with higher transparency).}
    \label{fig:critical_scenario}
\end{figure}

\figref{criticality_analysis} shows that vehicles tend to merge onto the mainline quite close behind another vehicle with $50\%$ having a \gls{PET} lower than $4.0$ seconds and $25\%$ lower than $2.6$ seconds. For the \textit{into} type the \gls{PET} is uniformly distributed within $[5.0; 15.1]$ seconds and a mean of $9.7\pm2.9$ seconds. The gap between a vehicle merging in front of another vehicle is even closer than for the \textit{behind} class with $50\%$ having a \gls{PET} lower than $3.0$ seconds and $25\%$ lower than $2.0$ seconds. Noteworthy is the situation with a \gls{PET} of $0.08$ seconds and thus denotes a near-crash situation. This situation is depicted in \figref{critical_scenario} where the ego vehicle starts the lane-change maneuver and one second later (in transparent). In that scenario, a vehicle (in blue) merges onto the mainline which is occupied by a truck (in green). Although the \gls{PET} indicates that the scenario is critical, \figref{critical_scenario}  shows that the situation becomes uncritical since the gap between both vehicles increases.

\section{Conclusion}
\label{sec:conclusion}
A scenario-based validation approach for \gls{CAV} demands information from real-world driving as provided by the \gls{TFNDS}. However, a challenge is to annotate the data with the scenarios of interest.

This work addresses this issue and proposes a novel methodology for on-ramp scenario extraction and analysis. The extracted scenarios are categorized and assessed using the \gls{PET} which also enables the identification of critical scenarios. Evaluations have shown that the approach robustly extracts scenarios with an accuracy of $95.14\%$. However, the results also indicate that the method fails identifying maneuvers starting at the merging lane's end. This will be the subject of future works.

Furthermore, the maneuver behavior in terms of the start and end position is analyzed. The first results indicate that most vehicles tend to merge on the mainline at the first half of the acceleration line. In follow-up works, a more comprehensive dataset will be employed to analyze further the behavior of the ego vehicle and challengers in on-ramp merging scenarios.

\IEEEtriggeratref{9}

\bibliographystyle{IEEEtran}
\bibliography{literature}

\begin{thebibliography}{10}
\providecommand{\url}[1]{#1}
\csname url@rmstyle\endcsname
\providecommand{\newblock}{\relax}
\providecommand{\bibinfo}[2]{#2}
\providecommand\BIBentrySTDinterwordspacing{\spaceskip=0pt\relax}
\providecommand\BIBentryALTinterwordstretchfactor{4}
\providecommand\BIBentryALTinterwordspacing{\spaceskip=\fontdimen2\font plus
\BIBentryALTinterwordstretchfactor\fontdimen3\font minus
  \fontdimen4\font\relax}
\providecommand\BIBforeignlanguage[2]{{%
\expandafter\ifx\csname l@#1\endcsname\relax
\typeout{** WARNING: IEEEtran.bst: No hyphenation pattern has been}%
\typeout{** loaded for the language `#1'. Using the pattern for}%
\typeout{** the default language instead.}%
\else
\language=\csname l@#1\endcsname
\fi
#2}}

\bibitem{gunther2019resource}
J.~G{\"u}nther, H.~Lehmann, P.~Nuss, and K.~Purr, \emph{Resource-efficient
  Pathways Towards Greenhouse-gas-neutrality-RESCUE: Summary Report}.\hskip 1em
  plus 0.5em minus 0.4em\relax Umweltbundesamt, 2019.

\bibitem{KOPELIAS2020135237}
P.~Kopelias, E.~Demiridi, K.~Vogiatzis, A.~Skabardonis, and V.~Zafiropoulou,
  ``{Connected \& autonomous vehicles – Environmental impacts – A
  review},'' \emph{Science of The Total Environment}, vol. 712, p. 135237,
  2020.

\bibitem{PAPADOULIS201912}
A.~Papadoulis, M.~Quddus, and M.~Imprialou, ``Evaluating the safety impact of
  connected and autonomous vehicles on motorways,'' \emph{Accident Analysis \&
  Prevention}, vol. 124, pp. 12--22, 2019.

\bibitem{7548168}
T.~{Guan} and C.~W. {Frey}, ``Predictive energy efficiency optimization of an
  electric vehicle using traffic light sequence information*,'' in \emph{2016
  IEEE International Conference on Vehicular Electronics and Safety (ICVES)},
  2016, pp. 1--6.

\bibitem{WinnerPegasus2018}
H.~Winner, K.~Lemmer, T.~Form, and J.~Mazzega, \emph{{PEGASUS---First Steps for
  the Safe Introduction of Automated Driving}}.\hskip 1em plus 0.5em minus
  0.4em\relax Switzerland: Springer International Publishing, 2018, ch. Vehicle
  Systems and Technologies Development, pp. 185--195.

\bibitem{2018Hallerbach}
S.~Hallerbach, Y.~Xia, U.~Eberle, and F.~K{\"o}ster, ``Simulation-based
  identification of critical scenarios for cooperative and automated
  vehicles,'' \emph{SAE Intl. J CAV}, vol.~1, pp. 93--106, 04 2018.

\bibitem{9330510}
C.~{Neurohr}, L.~{Westhofen}, M.~{Butz}, M.~H. {Bollmann}, U.~{Eberle}, and
  R.~{Galbas}, ``Criticality analysis for the verification and validation of
  automated vehicles,'' \emph{IEEE Access}, vol.~9, pp. 18\,016--18\,041, 2021.

\bibitem{damm2018formal}
W.~Damm, E.~M{\"o}hlmann, T.~Peikenkamp, and A.~Rakow, ``A formal semantics for
  traffic sequence charts,'' in \emph{Principles of Modeling}.\hskip 1em plus
  0.5em minus 0.4em\relax Springer, 2018, pp. 182--205.

\bibitem{9069432}
W.~{Qi}, W.~{Wang}, B.~{Shen}, and J.~{Wu}, ``A modified post encroachment time
  model of urban road merging area based on lane-change characteristics,''
  \emph{IEEE Access}, vol.~8, pp. 72\,835--72\,846, 2020.

\bibitem{elrofai2018streetwise}
H.~Elrofai, J.~Paardekooper, E.~d. Gelder, S.~Kalisvaart, and O.~Op~den Camp,
  ``Streetwise: scenario-based safety validation of connected automated
  driving,'' 2018.

\bibitem{Kumar_SVM_2013}
P.~{Kumar}, M.~{Perrollaz}, S.~{Lefèvre}, and C.~{Laugier}, ``Learning-based
  approach for online lane change intention prediction,'' in \emph{2013 IEEE
  Intelligent Vehicles Symposium (IV)}, 2013, pp. 797--802.

\bibitem{SVm_RBF}
R.~{Izquierdo}, I.~{Parra}, J.~{Muñoz-Bulnes}, D.~{Fernández-Llorca}, and
  M.~A. {Sotelo}, ``{Vehicle trajectory and lane change prediction using ANN
  and SVM classifiers},'' in \emph{2017 IEEE 20th International Conference on
  Intelligent Transportation Systems (ITSC)}, 2017, pp. 1--6.

\bibitem{Ramyar_OneClassSVM}
S.~{Ramyar}, A.~{Homaifar}, A.~{Karimoddini}, and E.~{Tunstel},
  ``{Identification of anomalies in lane change behavior using one-class
  SVM},'' in \emph{2016 IEEE International Conference on Systems, Man, and
  Cybernetics (SMC)}, Oct 2016, pp. 004\,405--004\,410.

\bibitem{2018MonotANN}
N.~{Monot}, X.~{Moreau}, A.~{Benine-Neto}, A.~{Rizzo}, and F.~{Aioun},
  ``{Comparison of rule-based and machine learning methods for lane change
  detection},'' in \emph{2018 21st International Conference on Intelligent
  Transportation Systems (ITSC)}, Nov 2018, pp. 198--203.

\bibitem{2019Erdogan}
A.~{Erdogan}, B.~{Ugranli}, E.~{Adali}, A.~{Sentas}, E.~{Mungan}, E.~{Kaplan},
  and A.~{Leitner}, ``{Real-World Maneuver Extraction for Autonomous Vehicle
  Validation: A Comparative Study},'' in \emph{2019 IEEE Intelligent Vehicles
  Symposium (IV)}, June 2019, pp. 267--272.

\bibitem{9304560}
F.~{Montanari}, R.~{German}, and A.~{Djanatliev}, ``Pattern recognition for
  driving scenario detection in real driving data,'' in \emph{2020 IEEE
  Intelligent Vehicles Symposium (IV)}, 2020, pp. 590--597.

\bibitem{8569682}
F.~{Kruber}, J.~{Wurst}, and M.~{Botsch}, ``{An Unsupervised Random Forest
  Clustering Technique for Automatic Traffic Scenario Categorization},'' in
  \emph{2018 21st International Conference on Intelligent Transportation
  Systems (ITSC)}, Nov 2018, pp. 2811--2818.

\bibitem{klitzke_itsc2020}
L.~{Klitzke}, C.~{Koch}, and F.~{Köster}, ``{Identification of Lane-Change
  Maneuvers in Real-World Drivings With Hidden Markov Model and Dynamic Time
  Warping},'' in \emph{2020 IEEE 23rd International Conference on Intelligent
  Transportation Systems (ITSC)}, 2020, pp. 1--7.

\bibitem{8506402}
W.~{Wang}, J.~{Xi}, and D.~{Zhao}, ``{Driving Style Analysis Using Primitive
  Driving Patterns With Bayesian Nonparametric Approaches},'' \emph{IEEE
  Transactions on Intelligent Transportation Systems}, vol.~20, no.~8, pp.
  2986--2998, Aug 2019.

\bibitem{9272025}
P.~{Elspas}, J.~{Langner}, M.~{Aydinbas}, J.~{Bach}, and E.~{Sax}, ``Leveraging
  regular expressions for flexible scenario detection in recorded driving
  data,'' in \emph{2020 IEEE International Symposium on Systems Engineering
  (ISSE)}, 2020, pp. 1--8.

\bibitem{18626}
L.~Rabiner, ``A tutorial on hidden markov models and selected applications in
  speech recognition,'' \emph{Proceedings of the IEEE}, vol.~77, no.~2, pp.
  257--286, 1989.

\bibitem{bilmes1998gentle}
J.~A. Bilmes \emph{et~al.}, ``{A gentle tutorial of the EM algorithm and its
  application to parameter estimation for Gaussian mixture and hidden Markov
  models},'' \emph{International Computer Science Institute}, vol.~4, no. 510,
  p. 126, 1998.

\bibitem{behbahani2015framework}
H.~Behbahani and N.~Nadimi, ``A framework for applying surrogate safety
  measures for sideswipe conflicts,'' \emph{International Journal for Traffic
  \& Transport Engineering}, vol.~5, no.~4, pp. 371--383, 2015.

\bibitem{2003douglas}
D.~Gettman and L.~Head, ``Surrogate safety measures from traffic simulation
  models,'' \emph{Transportation Research Record}, vol. 1840, no.~1, pp.
  104--115, 2003.

\bibitem{2019weber}
H.~Weber, J.~Bock, J.~Klimke, C.~Roesener, J.~Hiller, R.~Krajewski, A.~Zlocki,
  and L.~Eckstein, ``A framework for definition of logical scenarios for safety
  assurance of automated driving,'' \emph{Traffic Injury Prevention}, vol.~20,
  no. sup1, pp. S65--S70, 2019.

\bibitem{2017_carla}
A.~Dosovitskiy, G.~Ros, F.~Codevilla, A.~Lopez, and V.~Koltun, ``{CARLA}: {An}
  open urban driving simulator,'' in \emph{Proceedings of the 1st Annual
  Conference on Robot Learning}, ser. Proceedings of Machine Learning Research,
  S.~Levine, V.~Vanhoucke, and K.~Goldberg, Eds., vol.~78.\hskip 1em plus 0.5em
  minus 0.4em\relax PMLR, 13--15 Nov 2017, pp. 1--16.

\bibitem{dlr71460}
M.~Behrisch, L.~Bieker, J.~Erdmann, and D.~Krajzewicz, ``{SUMO {--} Simulation
  of Urban MObility: An Overview},'' in \emph{SIMUL 2011}, S.~.~U. of~Oslo
  Aida~Omerovic, R.~I. R. T. P. D.~A. Simoni, and R.~I. R. T. P.~G. Bobashev,
  Eds.\hskip 1em plus 0.5em minus 0.4em\relax ThinkMind, Oktober 2011.

\bibitem{5509799}
M.~Werling, J.~Ziegler, S.~Kammel, and S.~Thrun, ``Optimal trajectory
  generation for dynamic street scenarios in a frenét frame,'' in \emph{2010
  IEEE International Conference on Robotics and Automation}, 2010, pp.
  987--993.

\end{thebibliography}
\end{document}